%% file: tmi.tex
\documentclass[journal,twoside,web]{ieeecolor}
\usepackage{tmi}
\usepackage{cite}
\usepackage{amsmath,amssymb,amsfonts}
\usepackage{algorithmic}
\usepackage{graphicx}
\usepackage{textcomp}
\usepackage{booktabs}
\usepackage{multirow}
\usepackage{adjustbox}
\usepackage[table]{xcolor}
\usepackage{xcolor}
\usepackage{capt-of}

\newcommand{\eg}{\textit{e}.\textit{g}.}
\definecolor{titlecolor}{RGB}{238,248,238} 
\definecolor{highlight}{RGB}{235,248,250}     
\usepackage[colorlinks,hyperindex,breaklinks]{hyperref}
\def\BibTeX{{\rm B\kern-.05em{\sc i\kern-.025em b}\kern-.08em
    T\kern-.1667em\lower.7ex\hbox{E}\kern-.125emX}}
\begin{document}
\title{Token Merging via Spatiotemporal Information Mining for Surgical Video Understanding}


\author{Xixi Jiang, Chen Yang, Dong Zhang, Pingcheng Dong, Xin Yang, Kwang-Ting Cheng, \IEEEmembership{Fellow, IEEE}
\thanks{This work was supported in part by the Hong Kong SAR RGC General Research Fund under Grant 16208823, the National Key R\&D Program of China (2024YFE0217700), National Natural Science Foundation of China (62472184), and the Fundamental Research Funds for the Central Universities.}
\thanks{X. Jiang, C. Yang, D. Zhang, P. Dong, and K.-T. Cheng are with the Department of Electronic and Computer Engineering, The Hong Kong University of Science and Technology, Hong Kong, China. E-mail: xjiangbh@connect.ust.hk, eechengyang@ust.hk, dongz@ust.hk, pingcheng.dong@connect.ust.hk, timcheng@ust.hk}
\thanks{X. Yang is with the School of Electronic Information and Communications, Huazhong University of Science and Technology, Wuhan, China. E-mail: xinyang2014@hust.edu.cn.}
\thanks{Corresponding author: Kwang-Ting Cheng.}}

\maketitle

\input{sections/0_abstract}

\input{sections/1_introduction}
\input{sections/2_relatedwork}

\input{sections/3_methods}

\input{sections/4_experiments}

\input{sections/5_discussion}

\bibliographystyle{IEEEtran}
\bibliography{tmi}
\end{document}

%% file: sections/0_abstract.tex
\begin{abstract}
Vision Transformer models have shown impressive effectiveness in the surgical video understanding tasks through long-range dependency modeling. However, current methods suffer from prohibitive computational costs due to processing massive spatiotemporal tokens across video frames. While prior work on token merging has advanced model efficiency, they fail to adequately consider the inherent spatiotemporal structure of video data and overlook the heterogeneous nature of information distribution, leading to suboptimal performance. In this paper, we propose a spatiotemporal information mining token merging (STIM-TM) method, representing the first dedicated approach for surgical video understanding. STIM-TM introduces a decoupled strategy that reduces token redundancy along temporal and spatial dimensions independently. Specifically, the temporal component merges spatially corresponding tokens from consecutive frames using saliency weighting, preserving critical sequential information and maintaining continuity. Meanwhile, the spatial component prioritizes merging static tokens through temporal stability analysis, protecting dynamic regions containing essential surgical information. Operating in a training-free manner, STIM-TM achieves significant efficiency gains with over $65\%$ GFLOPs reduction while preserving competitive accuracy across comprehensive surgical video tasks. Our method also supports efficient training of long-sequence surgical videos, addressing computational bottlenecks in surgical applications. Code is available at https://github.com/xjiangmed/STIM-TM.
\end{abstract}


%% file: sections/1_introduction.tex
\section{Introduction}
\label{sec:introduction}

\begin{figure}[!t]
\centering{\includegraphics[width=0.48
\textwidth]{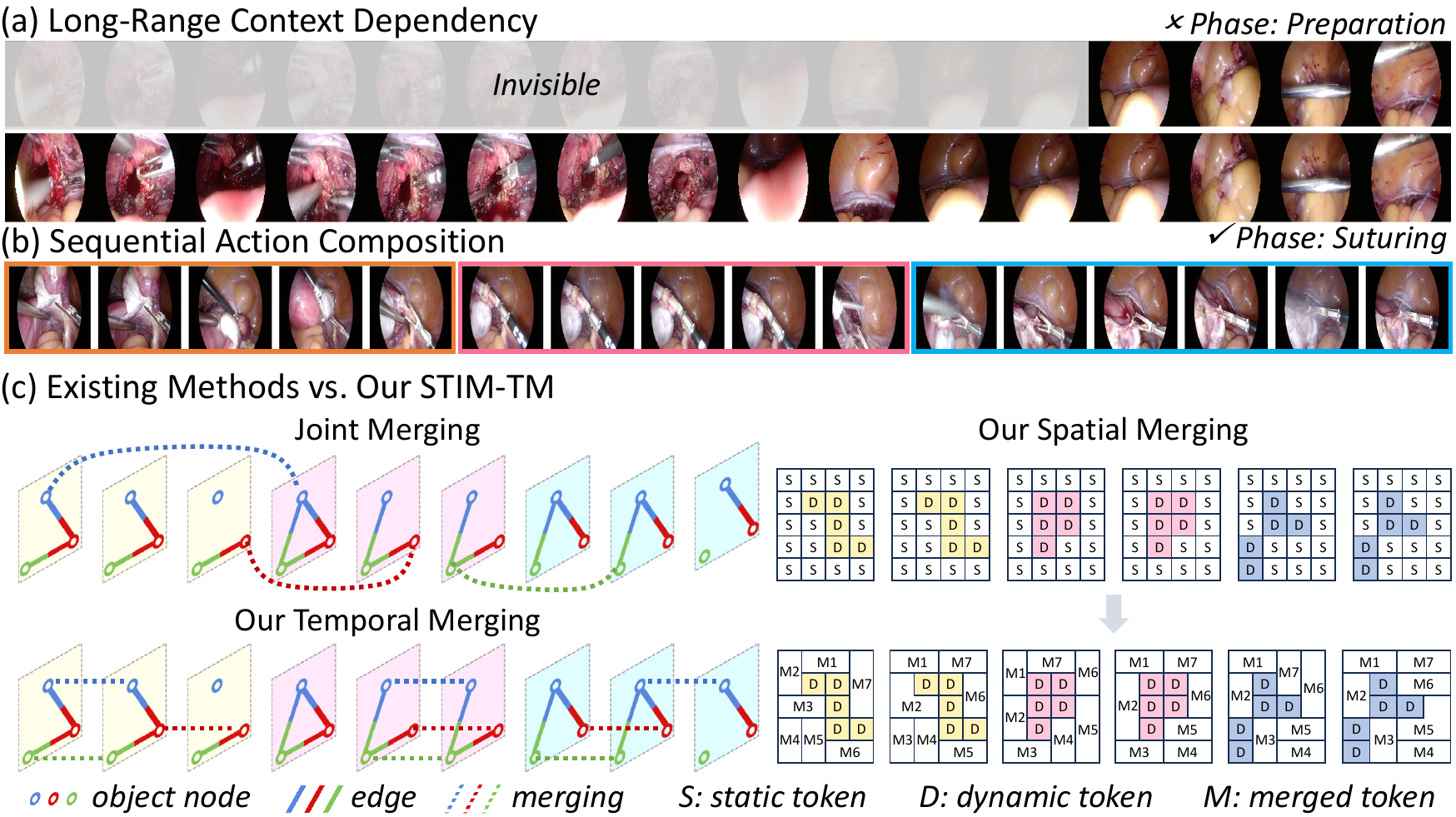}}
\caption{Characteristics of surgical videos and illustration of STIM-TM. (a) Long-term temporal context is crucial as short clips struggle to distinguish similar frames across phases, causing misclassifications such as confusing ``suturing" with ``preparation". (b) Surgical videos exhibit multi-scale sequential actions with strong interdependencies between actions. (c) Existing methods (\eg, joint merging) may merge similar objects across distant frames, disrupting action continuity (the left in (c) shows object-centric graph representation with cross-frame merging). Our STIM-TM decouples spatiotemporal merging: temporal merging operates between adjacent frames to preserve action sequences, while spatial merging (the right in (c)) retains more tokens in dynamic regions.}
\vspace{-6mm}
\label{fig:surgical_characteristic} 
\end{figure}

Surgical video understanding aims to extract meaningful insights about instrument interactions, anatomical structures, and procedural dynamics~\cite{khan2025surgical}. This task plays a pivotal role in advancing surgical practice, encompassing tasks ranging from coarse-level workflow recognition (\eg, phase recognition~\cite{LoViT,yang2024surgformer}, step recognition~\cite{ayobi2024pixel}) to fine-grained scene understanding (\eg, instrument segmentation~\cite{liu2025resurgsam2}, action triplet detection~\cite{pei2025instrument}). Thanks to the self-attention mechanism's ability to capture long-range dependencies, Vision Transformers (ViTs)\cite{dosovitskiy2020image} have been increasingly adopted for surgical video understanding, demonstrating remarkable effectiveness~\cite{yang2024surgformer,pei2025instrument}. However, ViT-based methods suffer from a severe computational bottleneck as self-attention complexity scales quadratically with the number of tokens~\cite{bolyatoken}. This challenge is particularly severe for video data, and becomes even more pronounced for high-resolution and long-duration surgical videos~\cite{LoViT,wang2025improving}. This bottleneck impedes the real-time deployment in resource-limited clinical environments.

To address quadratic complexity within ViT models, reducing computational burden of attention mechanisms has become a primary solution~\cite{liu2022videoswin,yang2024surgformer,LoViT}. Prior research has developed strategies such as decoupling self-attention across temporal and spatial dimensions~\cite{timesformer21}, or restricting attention to local spatiotemporal windows~\cite{liu2022videoswin}. Surgical-specific optimizations include hierarchical two-stage temporal learning~\cite{SKiT,LoViT}, sparse sampling~\cite{yang2024surgformer}, and Mamba-based architectures~\cite{liu2025resurgsam2,pei2025instrument}. While effective, these approaches typically require architectural modifications and training from scratch. In contrast, token merging~\cite{bolyatoken,norouzi2024algm,lee2024multi} methods offer a flexible, training-free alternative which reduce sequence length by merging redundant tokens, decreasing computational overhead. The representative work ToMe~\cite{bolyatoken} proposes Bipartite Soft Matching (BSM) for efficiently merging highly similar tokens. Subsequent studies~\cite{norouzi2024algm,lee2024multi,yang2025visionzip,shang2025prumerge,tao2024dycoke} have built upon BSM with various improvements across image and video domains. 

However, existing video token merging methods apply uniform merging strategies across all tokens and dimensions, without considering dimensional context or semantic importance. These uniform strategies lead to two critical limitations. First, they ignore the inherent spatiotemporal structure of video data, risking disruption of critical semantic relationships. Second, they fail to account for varying information density across different visual regions, potentially degrading the discriminative information of merged tokens. To achieve efficient token merging, two fundamental challenges should be addressed. The first one is: \textbf{\emph{How should tokens be merged to preserve the distinct characteristics of spatial and temporal dimensions?}} Existing unidimensional methods eliminate either spatial redundancy~\cite{yang2025visionzip,shang2025prumerge} within individual frames or temporal redundancy~\cite{tao2024dycoke} across sequences, resulting in asymmetric compression and information imbalance across dimensions. Joint merging methods~\cite{bolyatoken,choi2024vid,lee2024video,shen2024tempme} simultaneously consider both dimensions but uniformly process all tokens without dimensional distinction. However, as shown in Fig.~\ref{fig:surgical_characteristic} (a-b), to extract discriminative features from surgical videos, different dimensions have distinct requirements: the temporal dimension requires preserving long-term context and action sequence integrity, while the spatial dimension requires retaining local structure to preserve distinctive action characteristics. Unfortunately, as illustrated in Fig.~\ref{fig:surgical_characteristic} (c) (left), joint merging may merge similar objects across distant frames, disrupting temporal sequence coherence while compromising spatial feature preservation. Since temporal adjacency and spatial locality carry different semantic meanings in surgical videos, this necessitates a decoupled framework that leverages dimension-specific characteristics.

The second one is: \textbf{\emph{Where should tokens be merged to maximize redundancy reduction while preserving discriminative information?}} We argue that merging decisions should consider information density and leverage the spatiotemporal structure of video. Our analysis in Sec.~\ref{sec:analysis} reveals two video-specific redundancy patterns: (1) \textbf{Temporal Redundancy Pattern}:  \emph{Spatially corresponding tokens between adjacent frames exhibit significantly higher similarities than token pairs between non-adjacent frames or at different spatial positions}. This reflects procedural continuity in surgical videos, indicating that temporal merging should be constrained to spatially aligned tokens from adjacent frames to reduce redundancy effectively. (2) \textbf{Spatial Redundancy Pattern}: Within individual frames, information distribution is highly non-uniform. \emph{Static regions (\emph{e.g.}, undisturbed peripheral tissues) exhibit consistent features, while dynamic regions (e.g., moving instruments, deforming tissues) show significant variation}. This suggests that spatial merging should target low-entropy static regions while preserving informative dynamic elements.

Based on these observations, we propose SpatioTemporal Information Mining Token Merging (STIM-TM), a novel training-free module for video transformers. From an information bottleneck perspective, STIM-TM strategically selects merging candidates to maximize task-relevant information retention while minimizing input redundancy. Specifically, as illustrated in Fig.~\ref{fig:surgical_characteristic} (c), STIM-TM decouples the merging process into independent temporal and spatial dimensions, preserving both temporal continuity and spatial structure. For temporal merging, we restrict merging candidates to spatially corresponding positions across adjacent frames, and iteratively merge the most similar token pairs. Temporal attention scores serve as saliency measures to weight the merge operation, ensuring more informative tokens contribute more to the merged result. For spatial merging, we employ segment-based consistency scores to prioritize merging static regions while protecting dynamic objects. The main contributions are summarized as follows:

\begin{figure*}[!t]
\centering{\includegraphics[width=0.84\textwidth]
{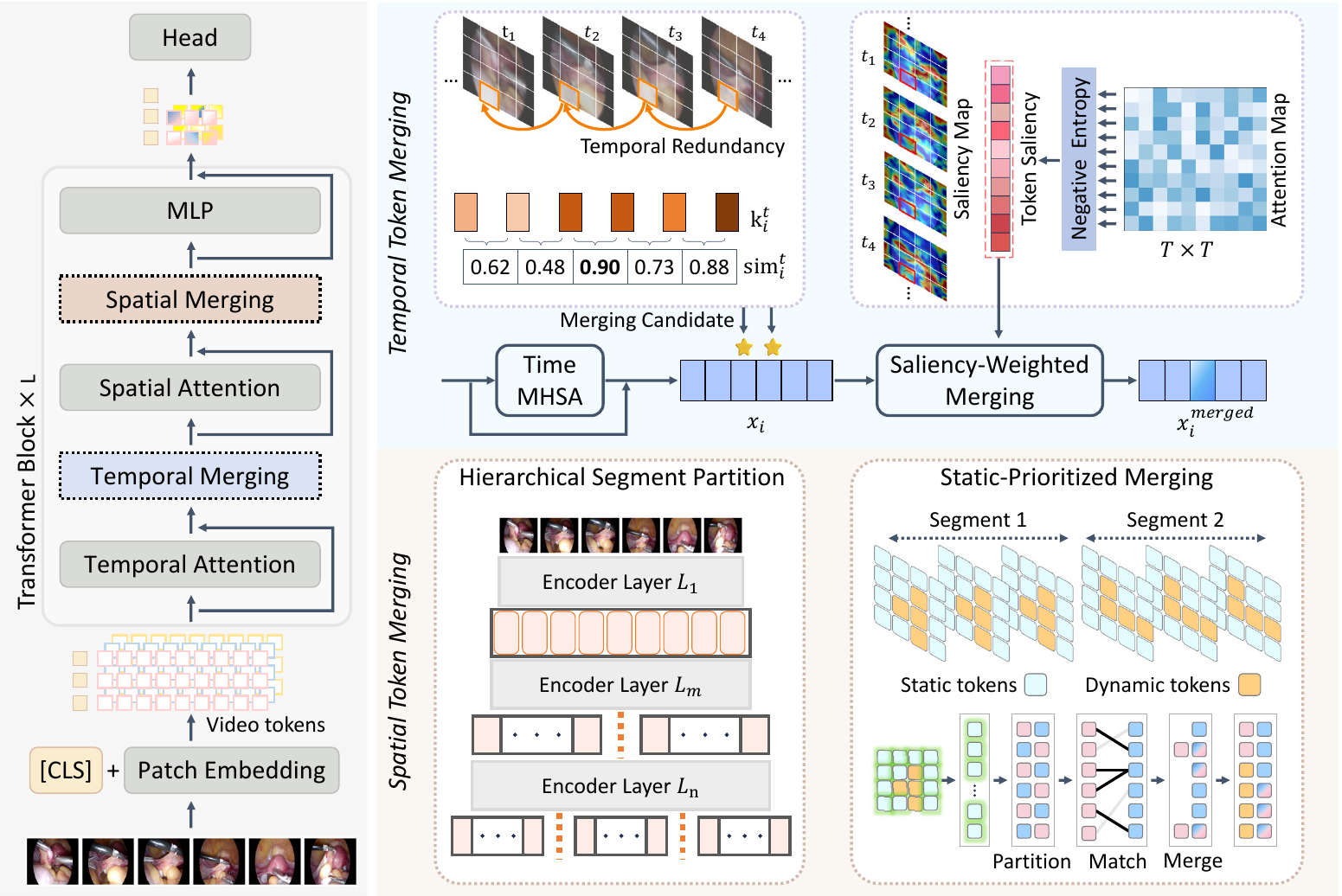}}
\caption{(Left) Baseline architecture and the insertion position of STIM-TM. STIM-TM is a plug-and-play module that reduces temporal and spatial redundancy independently. (Upper right) Temporal merging: We calculate temporal redundancy for each spatial position based on adjacent-frame similarity. The most similar adjacent frames are identified as candidates and aggregated using saliency weights from attention maps. (Lower right) Spatial merging: Video features are hierarchically partitioned into temporal segments across encoder layers. Within each segment, spatial positions are assigned static scores based on cross-frame similarity. High static-score tokens are prioritized and merged via bipartite soft matching.}\label{fig:framework}
\end{figure*}

\begin{itemize}
\item We introduce STIM-TM, the first dedicated token merging framework for surgical video transformers, leveraging the unique spatiotemporal redundancy patterns.
\item We demonstrate the broad applicability of STIM-TM by achieving significant computational reduction with minimal accuracy degradation across diverse tasks.
\item STIM-TM is a plug-and-play module that can be seamlessly integrated into existing video transformer architectures during inference, or used during training to enable long-sequence modeling under memory constraints.
\end{itemize}

%% file: sections/2_relatedwork.tex
\section{Related Work}
\label{sec:related work}
\subsection{Surgical Video Understanding with Vision Transformer}

Vision Transformers (ViTs)~\cite{dosovitskiy2020image} have recently outperformed traditional CNNs in surgical video understanding by effectively modeling complex spatiotemporal dependencies. LoViT~\cite{LoViT} and SKiT~\cite{SKiT} employ two-stage, fully transformer-based architecture with spatial feature extraction followed by temporal transformers for long-range dependency modeling in phase recognition. Surgformer~\cite{yang2024surgformer} improves phase recognition using hierarchical temporal attention to capture local and global dynamics. TAPIS~\cite{ayobi2024pixel} extends transformer architectures to broader workflow analysis, integrating a video-wise encoder with a region proposal network for multiple tasks. With the advent of large-scale surgical video datasets, transformer-based video foundation models~\cite{wang2023foundation,wang2025improving,jaspers2025scaling} have emerged as a transformative paradigm for surgical video understanding, exhibiting exceptional robustness and cross-domain generalization capabilities. 
Nevertheless, the high computational cost of transformer-based approaches remains a significant challenge. To address this, our STIM-TM merges redundant tokens across spatial and temporal dimensions to decrease the token length of transformer blocks, significantly reducing computational cost.

\subsection{Token Reduction}
Token reduction methods decrease computational overhead through token pruning and merging. \textbf{Token pruning} methods~\cite{rao2021dynamicvit,wang2022efficient,ding2023prune} dynamically remove uninformative tokens based on attention scores or learnable modules. Recent works such as STTS~\cite{wang2022efficient} and STA~\cite{ding2023prune} extend pruning to video tasks. However, most pruning methods require additional training, and aggressive pruning may discard critical features. \textbf{Token merging} has emerged as an effective, often training-free method for reducing token redundancy. Among these methods, ToMe~\cite{bolyatoken} pioneers this direction by merging semantically similar tokens via Bipartite Soft Matching. Subsequent works~\cite{norouzi2024algm,lee2024multi} have explored various merging strategies, such as local-then-global merging~\cite{norouzi2024algm} and multi-criteria merging~\cite{lee2024multi}. While these methods are effective, most focus on spatial redundancy in images. Recently, token merging has been applied to video tasks~\cite{choi2024vid,ren2023testa,tao2024dycoke,yang2025visionzip, shang2025prumerge,li2024vidtome} to address temporal redundancy in long video understanding and diffusion models. VidToMe~\cite{li2024vidtome} improves temporal consistency in video generation by matching and merging tokens across frames within the self-attention module. VTM~\cite{lee2024video} introduces a learnable merging method that dynamically merges tokens based on their saliency. vid-TLDR~\cite{choi2024vid} uses attention score sharpness for saliency-aware merging. TESTA~\cite{ren2023testa} performs token merging along temporal and spatial dimensions, aggregating similar frames and tokens. TempMe~\cite{shen2024tempme} progressively merges tokens across neighboring clips. STPM~\cite{su2024stpm} integrates pruning and merging with semantic importance and attention-based strategies. Despite these advances, most methods inadequately handle spatiotemporal characteristics and overlook video-specific information patterns. STIM-TM overcomes these limitations by identifying low-information tokens across temporal and spatial dimensions, enabling targeted merging that preserves key structural information.

%% file: sections/3_methods.tex
\section{Methodology}
\label{sec:methodology}

\subsection{Preliminaries}
\noindent\textbf{Video Transformers.} Video transformers extend the standard transformer architecture~\cite{dosovitskiy2020image} to spatiotemporal data. The core self-attention mechanism remains:
\begin{equation}
    \text{Attention}(Q, K, V) = \text{softmax}\left(\frac{QK^\top}{\sqrt{C}}\right)V, 
\end{equation}
where $Q,K,V \in \mathbb{R}^{N \times C}$ are projections of input $X$. A $T$-frame clip ($H \times W$ resolution) is tokenized into $N = \frac{T}{t} \times \frac{H}{P} \times \frac{W}{P}$ spatiotemporal tokens. All tokens interact through either: (1) Joint space-time attention~\cite{dosovitskiy2020image} with full cross-token interactions. (2) Divided space-time attention~\cite{timesformer21} that sequentially computes in two dimensions: $\text{Attention}(Q_t, K_t, V_t)$ along temporal dimension and $\text{Attention}(Q_s, K_s, V_s)$ within each frame. The divided approach reduces complexity from $\mathcal{O}(N^2)$ to $\mathcal{O}(N_t \cdot N_s^2 + N_s \cdot N_t^2)$ where $N_t$ and $N_s$ are the number of temporal and spatial tokens respectively. However, both paradigms remain computationally prohibitive due to quadratic token scaling for long videos.

\noindent\textbf{Token Merging.} ToMe~\cite{bolyatoken} serves as the foundation for most subsequent token merging methods. It progressively reduces token redundancy by merging $R$ tokens at each transformer layer through Bipartite Soft Matching (BSM). BSM operates through three sequential steps: (1) Partitioning: Input tokens are alternately divided into two disjoint sets, $\mathcal A$ and $\mathcal B$. (2) Matching: Each source token in $\mathcal A$ is assigned to its most similar target token in $\mathcal B$, typically based on cosine similarity between their attention key representations. (3) Merging: The top-$R$ most similar matched pairs are merged via average pooling. This process occurs between the attention and MLP blocks, reducing the token count while preserving feature propagation. This plug-in approach is appealing for video transformers to enhance efficiency. However, ToMe overlooks the varying informativeness of spatiotemporal tokens.

\begin{figure}[!t]
\centering{\includegraphics[width=0.48\textwidth]{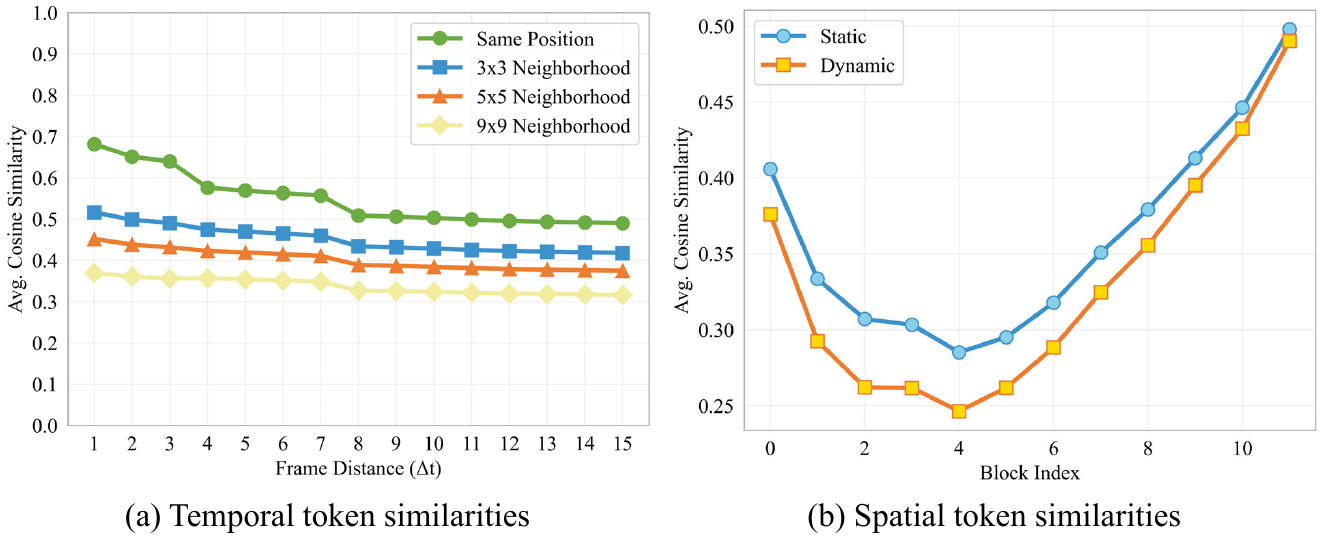}}
\caption{Token similarity analysis across temporal and spatial dimensions. (a) Temporal similarities at varying inter-frame distances on AutoLaparo~\cite{wang2022autolaparo} dataset (ninth transformer block). (b) Block-wise intra-frame similarity of static vs. dynamic tokens on CholecSeg8k~\cite{hong2020cholecseg8k} dataset.}\label{fig:simanalysis}
\end{figure}

\subsection{Theoretic Inspiration}
\label{sec:analysis}
\noindent\textbf{Token Similarity Analysis.} To identify optimal token merging candidates, we analyze features of Surgformer (trained on Autolaparo~\cite{wang2022autolaparo}) using cosine similarity. (1) Temporal Information Pattern. As shown in Fig.~\ref{fig:simanalysis}~(a), we compute similarity between: (i) tokens at identical spatial positions across varying inter-frame intervals, and (ii) tokens at different spatial positions within $W=k{\times}k$ neighborhood windows across equivalent temporal separations. We find that spatially corresponding tokens in adjacent frames exhibit the highest similarity. Following principles from representation learning theory~\cite{oord2018representation}, this suggests that such token pairs encode more shared information, consistent with higher mutual information: $I(x_t^i, x_{t+1}^i) > I(x_t^i, x_{t+d}^{W}), \forall d>1$. (2) Spatial Information Pattern. As shown in Fig.~\ref{fig:simanalysis}~(b), tokens within each frame are classified as static or dynamic based on their temporal consistency in segmentation masks (dynamic: the mask label changes between two consecutive frames). The analysis reveals that static tokens consistently have higher intra-frame similarity across transformer blocks, which suggests that static tokens exhibit lower conditional entropy given their spatial context, \emph{i.e.}, $H(X_{static}|Context) < H(X_{dynamic}|Context)$. This implies that static regions are highly predictable based on their surroundings, while dynamic regions contain more unpredictable and task-relevant information.

\noindent\textbf{Information Bottleneck Formulation.} Let $Z$ represent the original output tokens, $Z_{m}$ denote the merged tokens, $X$ be the input image, $Y$ be the task label. The information bottleneck (IB) principle~\cite{tishby2015deep} seeks to maximize the mutual information between $Z_{m}$ and $Y$ while minimizing the mutual information between $Z_{m}$ and $X$. In other words, effective token merging should maximize task-relevant information while minimizing input redundancy. We formulate STIM-TM as approximately optimizing: $min \ O_{IB} = I(Z_{m}, X) - I(Z_{m}, Y)$. Directly optimizing this objective faces challenges such as the non-separability of mutual information and computational complexity~\cite{wang2025efficient}. Our training-free approach implicitly follows the IB principle and decomposes $O_{IB}$ into temporal and spatial components: $O_{\text{STIM-TM}} = \alpha O_{\text{temporal}} + (1-\alpha) O_{\text{spatial}}$. Here, both temporal and spatial components contribute to the IB objective. $O_{\text{temporal}}$ minimizes $I(Z_{m}, X)$ by merging high mutual information tokens (spatially aligned adjacent frames) while maximizing $I(Z_{m}, Y)$ through saliency-weighted fusion. Meanwhile, $O_{\text{spatial}}$ minimizes $I(Z_{m}, X)$ by prioritizing merging static tokens while maximizing $I(Z_{m}, Y)$ by preserving more tokens in task-relevant dynamic regions. In practice, we employ computationally efficient heuristics to approximate the IB objectives. These practical implementations are detailed in the following sections. 


\subsection{Overview of STIM-TM}
As illustrated in Fig.\ref{fig:framework}, STIM-TM processes densely sampled input frames and progressively merges semantically similar tokens during encoding to reduce computational overhead.  Given an input video $V \in \mathbb{R}^{T \times H \times W \times 3}$, we tokenize each frame into $N_S$ patches and prepend a $[CLS]$ token for global video representation. Using TimeSformer~\cite{timesformer21} as our example encoder, we insert temporal token merging (TIM-TM) in Sec.~\ref{sec:3:4} after temporal attention and spatial token merging (SIM-TM) in Sec.~\ref{sec:3:5} after spatial attention. After each merging, TIM-TM reduces $R_T$ frames while SIM-TM reduces $R_S$ tokens per frame. These components can be deployed: either sequentially across different transformer blocks (\emph{e.g.}, temporal merging in earlier layers, spatial merging in later layers), or jointly within the same blocks (with temporal merging preceding spatial merging). Both strategies achieve effective token reduction with different efficiency-accuracy trade-offs, as detailed in the experimental section.

\subsection{Temporal Token Merging}
\label{sec:3:4}
Our analysis in Sec.~\ref{sec:analysis} reveals that spatially corresponding tokens between adjacent frames exhibit significantly higher similarities, suggesting that effective temporal merging should respect this temporal structure. However, existing methods apply uniform strategies without considering temporal proximity, potentially disrupting temporal coherence. To address this limitation, we propose adjacent-frame spatial-aligned merging that compresses consecutive frames while preserving essential procedural dynamics and temporal sequence integrity.
To achieve this goal, our temporal merging restricts merging candidates to consecutive frames and iteratively merges the most similar temporal pairs. We first calculate temporal redundancy as the similarity of features at the same spatial location across consecutive video frames. Formally, we compute the cosine similarity between the attention keys $\mathbf{k}_i^t$ and $\mathbf{k}_i^{t+1}$ at the $i$-th spatial location in frames $t$ and $t+1$:
\begin{equation}
\text{sim}_i^t = \frac{\mathbf{k}_i^t \cdot \mathbf{k}_i^{t+1}}{|\mathbf{k}_i^t|\, |\mathbf{k}_i^{t+1}|}.
\end{equation}
For each spatial position $i \in \{1, ..., N_S\}$ across $T$ frames, we compute similarities for all adjacent pairs $(t, t+1)$. Next, we identify the pair of adjacent frames $(\hat{t}, \hat{t}+1)$ with the highest similarity as merging candidates:
\begin{equation}
(\hat{t}, \hat{t}+1)=\arg\max_{t}(\text{sim}_i^t). 
\end{equation}
Finally, we merge the selected frame pair using saliency-weighted fusion to minimize task-relevant information loss. We compute the negative entropy of temporal attention scores $A_i^{t, t'}$ from the current transformer block as token saliency, which serves as an information-theoretic measure of token importance. For spatial token $i$ at time $t$, the negative entropy $H_i^{t}$ measures attention concentration:
\begin{equation}
H_i^{t} = \sum_{t'=1}^{T} A_i^{t, t'} \log A_i^{t, t'}.
\end{equation}
Higher $H_i^t$ values indicate focused attention (salient tokens, typically foreground actions), while lower values indicate dispersed attention (background regions). The saliency weights are obtained by normalizing the entropy values through min-max scaling:
\begin{equation}
\alpha_i^t = \frac{H_i^t - \min_{\tau}(H_i^\tau)}{\max_{\tau}(H_i^\tau) - \min_{\tau}(H_i^\tau)}.
\end{equation} 
We fuse tokens from the selected pair using these saliency weights:
\begin{equation}
x_i^{\text{merged}} = \alpha_i^{\hat{t}} * x_i^{\hat{t}} + \alpha_i^{\hat{t}+1} *x_i^{\hat{t}+1},
\end{equation}
where $\alpha_i^{\hat{t}}$ and $\alpha_i^{\hat{t}+1}$ are the weights of the selected frames $\hat{t}$ and $\hat{t}+1$ respectively. This entropy-based weighting embodies the information bottleneck principle by dynamically preserving high-information content during merging. 

We achieve $R_T$-frame reduction through $R_T$ successive merging iterations, merging the most similar adjacent pair at each step. This position-wise, adjacent-frame merging strategy preserves coherent spatial structures across compressed frames, without interfering with subsequent spatial modeling.

\subsection{Spatial Token Merging}
\label{sec:3:5}
Section~\ref{sec:analysis} highlights that spatial redundancy exhibits high non-uniformity: static regions have low information variance, while dynamic regions are rich in task-relevant content. To exploit this asymmetry, we propose static-prioritized merging, which aggressively compresses static areas and preserves more tokens for dynamic regions.
To identify stable static regions, we first partition $T$ frames into $K$ temporal segments. We compute the average cosine similarity $S(t)$ between corresponding spatial tokens in adjacent frames $t$ and $t+1$:
\begin{equation}
{S}(t) = \frac{1}{N_S}\sum_{i=1}^{N_S} \frac{\mathbf{k}_i^t \cdot \mathbf{k}_i^{t+1}}{|\mathbf{k}_i^t| |\mathbf{k}_i^{t+1}|},
\end{equation}
where $\mathbf{k}_i^t$ denotes the attention key of spatial token $i$ at time $t$. To detect temporal boundaries, we calculate a depth scoring~\cite{wang2024videollamb} function:
\begin{equation}
\text{D}(t) = \underset{\tau<t}{\max} S(\tau) + \underset{\tau>t}{\max} S(\tau) - 2S(t).
\end{equation}
Intuitively, $D(t)$ measures how distinct frame $t$ is from its neighbors; local maxima indicate transitions and are selected as segment boundaries. Specifically, we choose the $b=K-1$ largest local maxima of $D(t)$ as boundaries.
Within each temporal segment, we quantify the spatial stability of each spatial position $i$ by computing its static score $\mu_i$ based on cross-frame token similarity:
\begin{equation}
    \mu_i = \frac{2}{T_S(T_S-1)}\sum_{t=1}^{T_S}{\sum_{t'>t}{\frac{\mathbf{k}_i^t \cdot \mathbf{k}_i^{t'}}{|\mathbf{k}_i^t| |\mathbf{k}_i^{t'}|}}},
\end{equation}
where $T_S$ is the segment length and $\{\mathbf{k}_i^t\}_{t=1}^{T_S}$ denotes the token sequence for position $i$ across all frames in the segment. A higher $\mu_i$ indicates that position $i$ remains stable throughout the segment, while a lower value reflects dynamic behavior. 

We prioritize merging tokens with high static scores to retain informative regions. Specifically, for each merging operation that aims to remove $R_S$ tokens per frame, we first select the top $m \times R_S$ spatial positions with the highest static scores $\mu$ in a single frame as merging candidates. Following the ToMe framework, these candidates are partitioned into two disjoint sets, $\mathcal{A}$ and $\mathcal{B}$. Cross-set matching is then performed to identify the matched pairs, after which the top $R_S$ most similar pairs are merged via size-weighted averaging. This approach accounts for spatial information density by restricting merging to low-information stable regions, while preserving more unmerged tokens for dynamic areas.

\input{tables/ablation_token_merging}

To identify static regions at multiple temporal scales, temporal segmentation is performed at multiple granularity levels across encoder layers:
\begin{equation}
K = 
\begin{cases} 
  1 & \text{if } \ell \leq L/2 \\
  2 & \text{if } \ell > L/2, 
\end{cases}
\end{equation}
where $L$ is the total number of encoder layers, and $\ell$ is the layer index. Coarse segmentation in shallow layers identifies broadly stable regions across longer sequences, while finer segmentation in deeper layers captures more nuanced temporal variations. This hierarchical approach allows static redundancy of different granularities to be identified.

%% file: tables/ablation_token_merging.tex
\begin{table*}[t!]
\aboverulesep=0ex
\belowrulesep=0ex
\setlength{\tabcolsep}{3pt} 
\renewcommand{\arraystretch}{1.1} 
\footnotesize
\caption{Comparison of token reduction methods on Surgformer (Autolaparo dataset), including ablations on reduction methods, merging dimension, merging strategies, and order. The unrelaxed evaluation is used. T and S represent temporal and spatial merging. Symbols indicate insertion locations within the 12-layer transformer backbone: $\clubsuit$ T: Blocks 1-6, S: Blocks 7-12; $\dagger$ S: Blocks 1-6, T: Blocks 7-12; $\lozenge$ T: Blocks 1-6, S: Blocks 1-6.}

\centering
\begin{tabular}{l|l|c c | c c | c c c c}
    \toprule
     \multirow{2}{*}{Method} & \multirow{2}{*}{Type}
    & \multirow{2}{*}{$R_T$} 
    & \multirow{2}{*}{$R_S$} 
    & \multirow{2}{*}{GFLOPs $\downarrow$} 
    & \multirow{2}{*}{Memory(GB) $\downarrow$} 
    & \multicolumn{1}{c}{Video-level Metric}  
    & \multicolumn{3}{c}{Phase-level Metric}  \\
    \cmidrule(lr){7-7} \cmidrule(lr){8-10}
    & & & & & & Accuracy $\uparrow$ & Precision $\uparrow$ & Recall $\uparrow$ & Jaccard $\uparrow$ \\
    \midrule
    \multicolumn{2}{l|}{Baseline (w/o token reduction)}  & - & -  & 459.39 & 4.75 & 85.88 $\pm$ 5.82 &  83.45  & 74.33 &   65.13     \\  \midrule
    \rowcolor{titlecolor}
    \multicolumn{10}{l}{\textit{(1) Token Merging v.s. Pruning}} \\
    STTS~\cite{wang2022efficient} & Decoupled Pruning & 1 & 12 & 291.70 & 3.83 & 79.63 $\pm$ 7.31  &  72.65  & 67.02 & 56.83 \\
    STA~\cite{ding2023prune} & Joint Pruning & - & -  & 303.67 & 4.03 &  80.30 $\pm$ 4.52  & 78.52   & 69.22 & 55.71 \\
    DynamicViT~\cite{rao2021dynamicvit} & Decoupled Pruning & 1 & 12 & 300.37 & 3.12 &  84.82 $\pm$ 6.43  & 78.01    & 73.68 & 63.82 \\
    ToMe~\cite{bolyatoken} & Joint Merging & - & -& 304.47 & 3.78 &  80.99 $\pm$ 6.24  & 69.59   & 70.65 & 58.57 \\
    TempMe~\cite{shen2024tempme} & Joint Merging & - & - &289.21 & 3.66 & 84.09 $\pm$ 7.15 & 77.79 & 73.31 & 63.08 \\
    DyCoke~\cite{tao2024dycoke} & Temporal Merging & - & - & 314.26 & 2.85 & 81.64 $\pm$ 5.75 & 70.47 & 69.71 & 58.75 \\
    VisionZip~\cite{yang2025visionzip} & Spatial Merging & - & 12 & 292.61 & 3.54 & 84.52 $\pm$ 7.09 & 79.16 & 73.11 & 63.50 \\
    T-TESTA~\cite{ren2023testa} & Temporal Merging & 1 & -& 281.02 & 3.59 & 84.34 $\pm$ 5.67 & 76.59 & 73.92 & 62.99       \\
    S-TESTA~\cite{ren2023testa} & Spatial Merging & - & 12 & 292.64 & 3.66 & 84.36 $\pm$ 7.51 & 76.97 &  73.02 & 63.34      \\ 
    ST-TESTA~\cite{ren2023testa} & Decoupled Merging & 1 & 12  & 300.40 & 3.74 & 85.23 $\pm$ 6.08 & 78.44 & 74.59 & 64.69       \\
    \rowcolor{highlight}
    STIM-TM(Ours) & Decoupled Merging & 1 & 12 & 300.37 &  3.77 & 86.04 $\pm$ 5.58 & 80.47 & 74.32 & 65.20       \\
    \midrule
    \rowcolor{titlecolor}
    \multicolumn{10}{l}{\textit{(2) Merging dimension}} \\
     \multicolumn{2}{l|}{Only Temporal} & 1  & - & 281.02 & 3.70 & 85.67 $\pm$ 5.24   &  78.34  &  73.85 & 64.51   \\
     \multicolumn{2}{l|}{Only Spatial} & - & 12  & 299.61 & 3.55 & 85.96 $\pm$ 5.93   &  81.77  & 74.49 & 65.29  \\
     \multicolumn{2}{l|}{Both Temporal and Spatial} & 1 & 12 & 300.37 &  3.77 & 86.04 $\pm$ 5.58 & 80.47 & 74.32 & 65.20        \\
    \midrule
    \rowcolor{titlecolor}
    \multicolumn{10}{l}{\textit{(3) Temporal Merging Strategy}} \\
     \multicolumn{2}{l|}{Adjacent-Frame Merging} & 1 & -  & 281.02 & 3.70 &  85.47 $\pm$ 5.26 & 80.40   &   73.79 & 64.34     \\     
     \multicolumn{2}{l|}{w/ Saliency-Weighted Merging} & 1 & -  & 281.02 & 3.70 & 85.67 $\pm$ 5.24   &  78.34  &  73.85 & 64.51     \\
    \midrule
    \rowcolor{titlecolor} 
    \multicolumn{10}{l}{\textit{(4) Spatial Merging Strategy}} \\
    \multicolumn{2}{l|}{Intra-Frame Merging}& - & 12 & 292.64 & 3.66&  84.36 $\pm$ 7.51 & 76.97 &  73.02 & 63.34   \\     
     \multicolumn{2}{l|}{w/ Static-prioritized Merging (1 segment)} & - & 12 & 292.61 & 3.55 & 85.64 $\pm$ 5.79  &  80.62  &  74.28 & 64.86     \\
     \multicolumn{2}{l|}{w/ Static-prioritized Merging (2 segments)} & - & 12  & 292.60 & 3.55 & 85.38 $\pm$ 6.21  &  77.82  & 73.70  & 64.22    \\
     \multicolumn{2}{l|}{w/ Hierarchical Static-prioritized Merging}& - & 12  & 299.61 & 3.55 & 85.96 $\pm$ 5.93   &  81.77  & 74.49 & 65.29 \\
    \midrule
    \rowcolor{titlecolor}
    \multicolumn{10}{l}{\textit{(5) Merging Order}} \\
     \multicolumn{2}{l|}{Temporal First, then Spatial $\clubsuit$ } & 1 & 12 & 300.37 &  3.77 & 86.04 $\pm$ 5.58 & 80.47 & 74.32 & 65.20        \\
   \multicolumn{2}{l|}{Spatial First, then Temporal $\dagger$} & 1 & 12 & 304.05 & 3.72 & 85.13 $\pm$ 5.95   &  84.86  &  73.52 & 64.14     \\
     \multicolumn{2}{l|}{Parallel Temporal-Spatial $\lozenge$ } & 1 & 3 & 305.02 & 3.81 &  85.93 $\pm$ 5.64 & 78.44   & 73.99 &  64.80 \\
     \multicolumn{2}{l|}{Parallel Temporal-Spatial $\lozenge$} & 1 & 12 & 243.06 & 3.30 & 85.81 $\pm$ 5.95  & 82.20   & 74.49 & 65.09      \\
    \bottomrule
\end{tabular}
\label{tab:ablation_tokenreduction}
\end{table*}

%% file: sections/4_experiments.tex
\section{Experiment}
\label{sec:experiment}
\subsection{Experiment setup}
\noindent\textbf{Baselines and Datasets}. We apply STIM-TM as a training-free plug-in module to three strong surgical video understanding baselines: (1) \textbf{Surgformer}~\cite{yang2024surgformer}: A TimeSformer-based model for surgical phase recognition, evaluated on Cholec80~\cite{twinanda2016endonet} (80 cholecystectomy videos, 7 phases) and Autolaparo~\cite{wang2022autolaparo} (21 laparoscopic hysterectomy videos, 7 phases). We use temporal resolution of $T$ = 16 and a frame rate of 4 for both datasets. During testing, $T$ = 16 is used for Autolaparo, and $T$ = 24 is used for Cholec80. (2) \textbf{TAPIS}~\cite{ayobi2024pixel}: A hybrid framework fusing global spatiotemporal features from MViT\cite{fan2021multiscale} with instrument-aware regional features from Mask2Former~\cite{cheng2022masked}. We apply temporal-only token merging (TIM-TM) within MViT, as spatial pooling is inherent to its architecture. TAPIS model is trained on the GraSP~\cite{ayobi2024pixel} dataset (13 robot-assisted radical prostatectomy videos) using 16-frame clips. (3) \textbf{EndoFM-LV}~\cite{wang2025improving}: A self-supervised TimeSformer-based framework pre-trained on 6,469 endoscopy videos. We evaluate the effectiveness of STIM-TM on two downstream tasks: disease diagnosis using the PolypDiag dataset~\cite{tian2022contrastive} with fixed 32-frame clips, and polyp segmentation using the CVC-12K dataset~\cite{bernal2015wm} with variable-length sequences (6–25 frames). 

\noindent\textbf{Implementation details}. We follow the default settings of the three baselines~\cite{yang2024surgformer,ayobi2024pixel,wang2025improving} to train the models. All experiments are performed on NVIDIA RTX 3090 GPUs. For fair comparison, we adjust the merging number ($R_T$ and $R_S$) to ensure comparable computational overhead. For spatial merging, the hyperparameter $m$ is set to 2 for training-free inference, whereas it is set to 4 during model training. We evaluate the computational efficiency by measuring GFLOPs and Memory. To quantify IB performance, we compute the IB score $I(Z_{m}, X) - I(Z_{m}, Y)$ following LTM~\cite{wang2025efficient}, where the feature distributions of $Z_{m}$ and $X$ are estimated using class centroids and softmax-normalized distances. 
\input{tables/phase_recognition}

\input{tables/GraSP}

\subsection{Effectiveness on Different Baselines}
\noindent\textbf{Results on Surgical Phase Recognition.} As shown in Table~\ref{tab:ablation_tokenreduction} and Table~\ref{tab:Cholec80}, we evaluate token merging on Surgformer trained on Autolaparo and Cholec80 datasets, comparing baseline, TESTA, and STIM-TM. On Autolaparo (Table~\ref{tab:ablation_tokenreduction}), STIM-TM demonstrates significant advantages over TESTA while achieving performance comparable to the baseline at $R_T=1, R_S=12$ (Accuracy: 86.04\%; Jaccard: 65.20). On Cholec80 (Table~\ref{tab:Cholec80}), at $R_T=1$, $R_S=12$, STIM-TM exhibits minimal degradation (-0.37 relaxed accuracy, -0.6 relaxed Jaccard), while TESTA suffers substantial drops (-3.55 relaxed accuracy, -13.81 relaxed Jaccard). Notably, TESTA exhibits larger degradation at lower merging rates (13.81 relaxed Jaccard drop at $R_T=1$, $R_S=12$) compared to higher rates (5.65 drop at $R_T=2$, $R_S=24$). This reflects the instability of TESTA's global merging strategy, which may disrupt critical spatiotemporal structures. In contrast, STIM-TM maintains consistent superior performance, demonstrating exceptional preservation of discriminative features under aggressive reduction.

\noindent\textbf{Results on Holistic Surgical Scene Understanding.} Table~\ref{tab:GraSP} shows the performance comparison of token merging on TAPIS. Key observations reveal differential task sensitivity to information loss from merging: Instrument segmentation and action detection, which leverage spatial-temporal features indirectly via cross-attention mechanisms, exhibit significantly smaller performance degradation compared to phase recognition and step recognition. Critically, STIM-TM consistently outperforms TESTA across phase/step recognition and action detection, while performing comparably to TESTA on instrument segmentation. This demonstrates the robust adaptability and task-agnostic effectiveness of our proposed method.

\noindent\textbf{Results on Surgical Foundation Model.} Table~\ref{tab:EndoFM-LV-classification} and Table~\ref{tab:EndoFM-LV-segmentation} present the results on polyp classification and polyp segmentation tasks using EndoFM-LV, respectively. For the classification task, STIM-TM incurs no performance degradation at both merging rates. The segmentation task requires predicting dense, frame-wise segmentation maps, necessitating a token unmerging operation to restore the features' original resolution before inputting to the decoder. We implement token unmerging following the ALGM~\cite{norouzi2024algm} method, which replicates merged token embeddings at their original source token positions. As can be seen, STIM-TM consistently outperforms TESTA on both tasks. For the segmentation task, the performance is slightly reduced, with Dice decreases of 0.1 and 0.8 at the two settings, respectively. This demonstrates the broad applicability of our STIM-TM across different tasks.

\subsection{Extension to Training Stage}
STIM-TM is also applicable during training for reducing token length. We compare STIM-TM and TESTA across temporal (TIM-TM vs. T-TESTA), spatial (SIM-TM vs. S-TESTA), and combined dimensions (STIM-TM vs. ST-TESTA) on the Surgformer. In the temporal dimension, TIM-TM substantially outperforms TESTA. TESTA's global frame merging disrupts the procedural coherence, while our method preserves temporal continuity. In the spatial dimension, SIM-TM demonstrates superior performance over TESTA. TESTA’s spatial merging fails to consider the heterogeneous distribution of informative regions, while SIM-TM retains more task-relevant information. When combining temporal and spatial merging (applying parallel STIM-TM operations for optimal efficiency), STIM-TM achieves a 53\% reduction in GFLOPs and 64\% reduction in Memory, while maintaining minimal performance loss (0.1\% accuracy). Remarkably, TIM-TM alone exceeds baseline performance, suggesting that merging redundant tokens enables the model to concentrate on more informative features. 

\input{tables/foundationtable_mfig}

\input{tables/IB}

\subsection{Ablation Analysis}
We perform ablation studies on Surgformer trained on AutoLaparo to explore the answer of the following questions: (Q-1): Which token reduction strategy is more effective? (Q-2): Which dimension should be merged? (Q-3): Which temporal strategy performs better? (Q-4): Which spatial strategy performs better? (Q-5): What is the optimal merging order? (Q-6): How do merging rates affect performance?
\label{sec:ablation}

\noindent\textbf{Ans-1:~Results on Token Reduction Strategies.} We compare token pruning and merging methods in Table~\ref{tab:ablation_tokenreduction} (1). STTS~\cite{wang2022efficient} estimates the token importance through a learnable scoring network and requires fine-tuning, while other methods are training-free. Both pruning and merging methods reduce computation costs compared to the baseline. Most pruning methods (STTS, STA) perform worse than merging methods, indicating that token merging better retains valuable information. Decoupled merging (STIM-TM, TESTA) outperforms single-dimension merging (Dycoke, VisionZip) and joint merging (ToMe, TempMe), confirming that disentangled processing better preserves video structural information. STIM-TM achieves the best results among all methods. Compared with the baseline, STIM-TM maintains Accuracy and Jaccard while reducing computational overhead to 65.38\% GFLOPs and 79.36\% GPU memory. Table~\ref{tab:IB} shows the comparison of IB scores. Our STIM-TM delivers superior performance on both IB score and accuracy. Specifically, our method achieves the highest $I(Z_m,Y)$ for task-relevant information while maintaining reasonable $I(Z_m,X)$ for compression, resulting in the optimal IB score. This validates that our approach better approximates the IB objective by achieving superior balance between information compression and preservation.

\input{tables/sota_table_Autolaparo}
\input{tables/ablation_merging_rate}

\noindent\textbf{Ans-2:~Results on Merging Dimensions.}
In Table~\ref{tab:ablation_tokenreduction} (2), we compare token merging across different dimensions: temporal only, spatial only, and both. STIM-TM achieves superior performance when merging across both dimensions compared to single-dimension approaches. This aligns with TESTA's results in Table~\ref{tab:ablation_tokenreduction} (1). Single-dimension merging creates unbalanced compression, while dual-dimension merging achieves optimal results by integrating temporal and spatial compression.

\begin{figure*}[!t]
\centering{\includegraphics[width=\textwidth]{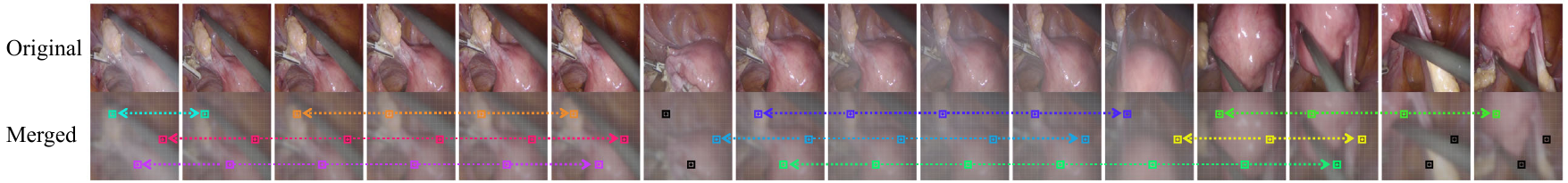}}
\caption{Visualization of temporal merging results from the 11th transformer layer of Surgformer, reducing the temporal dimension from 16 to 5 frames. We highlight merging results at three representative spatial locations, where different colors indicate distinct merged groups and black denotes unmerged tokens. Our TIM-TM preserves coherent temporal structure while retaining essential inter-frame dynamics.}\label{fig:T_merging}
\vspace{-3mm}
\end{figure*}

\begin{figure*}[!t]
\centering{\includegraphics[width=\textwidth]{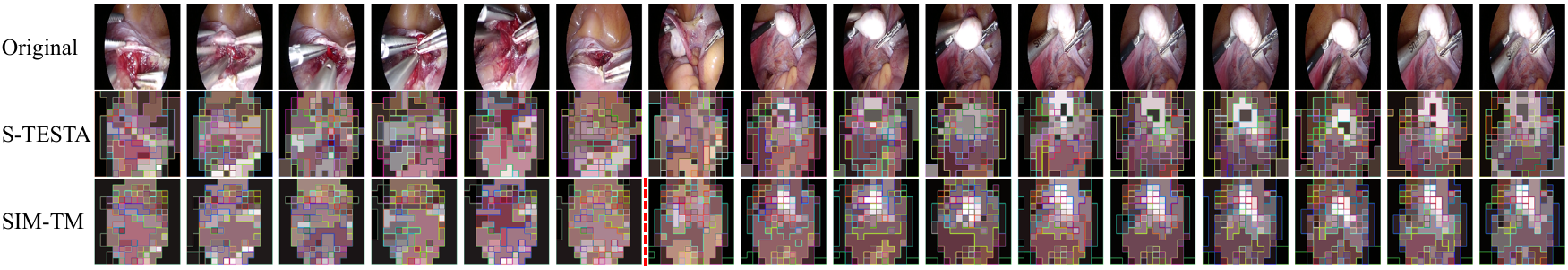}}
\caption{Spatial merging comparison between S-TESTA and SIM-TM at Surgformer's 12th transformer layer. Patches sharing identical interior and border colors represent merged groups. In the third row, the dotted line indicates the boundary between temporal segments. SIM-TM demonstrates segment-wise consistent merging and superior dynamic region preservation compared to S-TESTA's inconsistent frame-wise merging.}\label{fig:S_merging}
\vspace{-5mm}
\end{figure*}

\noindent\textbf{Ans-3:~Results on Temporal Merging Strategies.} We compare different temporal strategies in Table~\ref{tab:ablation_tokenreduction} (3). Compared with T-TESTA in Table~\ref{tab:ablation_tokenreduction} (1), which globally merges features of similar frames, our position-wise similar adjacent frame merging strategy achieves better performance, with an accuracy improvement of 1.13\%. Using saliency-weighted merging further improves the performance, indicating that saliency helps us preserve temporally critical features.

\noindent\textbf{Ans-4:~Results on Spatial Merging Strategies.} (1) Table~\ref{tab:ablation_tokenreduction} (4) compares different spatial strategies. We evaluate the impact of segment quantity $K$ through four configurations: intra-frame token merging (performing merging independently within each frame), single-segment ($K$=1), dual-segment ($K$=2), and hierarchical segmentation (1 segment in initial 6 blocks, 2 segments in final 6 blocks). Among all strategies, hierarchical segmentation achieves the best performance, outperforming intra-frame merging by 1.6\% accuracy improvement. This confirms that integrating global with local temporal context optimally identifies spatial redundancy. (2) In Fig.~\ref{fig:m_ablation}, we analyze the impact of merging candidate size $m$ on performance. We find that $m=2$ achieves optimal performance in training-free mode, while $m=4$ delivers the best results in training mode. For training-free, the frozen pre-trained model requires conservative merging of only the most certain static regions to avoid disrupting critical features, while for training, a larger candidate pool enables more flexible merging strategies.

\noindent\textbf{Ans-5:~Results on Merging Order.} We evaluated three merging orders (Table~\ref{tab:ablation_tokenreduction} (5)): (1) Temporal first, then Spatial, (2) Spatial first, then Temporal, and (3) Parallel Temporal-Spatial. Results revealed a performance hierarchy: Strategy (1) $>$ Strategy (3) $>$ Strategy (2). This ranking reflects the progressive nature of feature learning: early layers contain low-level spatial details, while deeper layers hold semantically abstracted representations with higher token similarities (as shown in Fig.~\ref{fig:simanalysis}). Strategy (1) achieves the best performance by aligning with this progression. Temporal merging first removes adjacent-frame redundancy while preserving spatial details, then spatial merging operates on abstracted features where merging is safer. Conversely, Strategy (2) prematurely discards vital spatial details, while Strategy (3)'s aggressive dual merging degrades both dimensions before sufficient feature extraction, despite its computational efficiency.

\noindent\textbf{Ans-6:~Results on Merging rates.} Table~\ref{tab:ablation_mergingrate} compares TESTA and STIM-TM across varying merging rates. Higher merging rates reduce computational cost but degrade performance. The configuration $R_T=1$, $R_S=12$ achieves optimal efficiency-accuracy trade-off. Further merging causes more severe loss of spatiotemporal information and greater performance drops. Notably, STIM-TM consistently outperforms TESTA at all merging rates, with its advantage becoming more pronounced as compression increases.

\subsection{Visualizations}
We visualize the temporal merging results on the AutoLaparo dataset in Fig.~\ref{fig:T_merging}. With $R_T=1$ per layer, temporal compression accumulates to reduce 16 frames to 5 by the 11th layer, yielding 5 groups per spatial position (colored highlights show merged groups and black tokens represent individual unmerged groups). The visualization shows that merged groups exhibit strong temporal clustering according to underlying surgical structures. At positions with significant inter-frame changes, tokens remain unmerged, ensuring temporally critical information is retained.
The results validate that TIM-TM successfully maintains temporal structural continuity and avoids disrupting essential temporal dependencies. 

We illustrate spatial merging with $R_S=12$ per layer in Fig.~\ref{fig:S_merging}, cumulatively reducing tokens from 196 to 52 per frame by layer 12. TESTA performs frame-independent merging without considering spatial information heterogeneity, leading to information loss in critical regions and inconsistent patterns across frames. SIM-TM distinguishes between static and dynamic tokens through cross-frame stability analysis, prioritizing the merging of static tokens. The visualization demonstrates SIM-TM's effectiveness: more tokens are allocated to dynamic regions requiring fine-grained details (\emph{e.g.}, surgical instruments), while reducing tokens in static areas. Additionally, SIM-TM maintains superior motion flow consistency and spatiotemporal structural coherence across frames.

%% file: tables/phase_recognition.tex
\begin{table*}[t]
\setlength{\tabcolsep}{3pt}
\renewcommand{\arraystretch}{1.1} 
\centering
\small
\caption{
Comparison of Token Merging Methods on Surgformer (Cholec80 dataset) with Different Merging Rates. Both relaxed and unrelaxed evaluations are used. We apply token merging to Temporal: Blocks 1-6, Spatial: Blocks 7-12.
}

\begin{tabular}{l|c c|c c|c|c c c c}
\toprule
\multirow{2}{*}{Evaluation}
    & \multirow{2}{*}{$R_T$}
    & \multirow{2}{*}{$R_S$}
    & \multirow{2}{*}{GFLOPs$\downarrow$}
    & \multirow{2}{*}{Memory(GB)$\downarrow$}
    & \multirow{2}{*}{Method}
    & \multicolumn{1}{c}{Video-level Metric}
    & \multicolumn{3}{c}{Phase-level Metric} \\
\cmidrule(lr){7-7} \cmidrule(lr){8-10}
    & & & & & & Accuracy$\uparrow$ & Precision$\uparrow$ & Recall $\uparrow$ & Jaccard $\uparrow$ \\
\midrule
\multirow{5}{*}{\parbox{1.2cm}{\centering Relaxed}}
    & - & - & 689.97 & 6.56 & Baseline & 93.33 $\pm$ 6.33 & 91.79 & 92.09 & 84.01 \\ 
    & \multirow{2}{*}{1} & \multirow{2}{*}{12} & 510.09 & 5.43 & ST-TESTA & 89.78 $\pm$ 6.10 & 84.79 & 82.71 & 70.20 \\
    
    & & & \cellcolor{highlight}510.05 & \cellcolor{highlight}5.47 & \cellcolor{highlight}STIM-TM(Ours) & \cellcolor{highlight}92.96 $\pm$ 6.70 & \cellcolor{highlight}91.97 & \cellcolor{highlight}91.28 & \cellcolor{highlight}83.41 \\ 
    & \multirow{2}{*}{2} & \multirow{2}{*}{24} & 366.68 & 4.30 & ST-TESTA & 91.64 $\pm$ 6.30 & 89.16 & 89.90 & 78.36 \\

    & & & \cellcolor{highlight}366.66 & \cellcolor{highlight}4.46 & \cellcolor{highlight}STIM-TM(Ours) & \cellcolor{highlight}92.35 $\pm$ 6.91 & \cellcolor{highlight}91.30 & \cellcolor{highlight}90.27 & \cellcolor{highlight}81.72 \\
\midrule
\multirow{5}{*}{\parbox{1.2cm}{\centering Unrelaxed}}
    & - & - & 689.97 & 6.56 & Baseline & 92.30 $\pm$ 6.31 & 87.82 & 89.27 & 79.79 \\
    & \multirow{2}{*}{1} & \multirow{2}{*}{12} & 510.09 & 5.43 & ST-TESTA & 88.11 $\pm$ 6.27 & 77.04 & 76.29 & 64.28 \\
    
    & & & \cellcolor{highlight}510.05 & \cellcolor{highlight}5.47 & \cellcolor{highlight}STIM-TM(Ours) & \cellcolor{highlight}91.97 $\pm$ 6.72 & \cellcolor{highlight}88.04 & \cellcolor{highlight}88.29 & \cellcolor{highlight}79.35 \\
    
    & \multirow{2}{*}{2} & \multirow{2}{*}{24} & 366.68 & 4.30 & ST-TESTA & 90.28 $\pm$ 6.31 & 83.42 & 85.12 & 73.27 \\
    
    & & & \cellcolor{highlight}366.66 & \cellcolor{highlight}4.46 & \cellcolor{highlight}STIM-TM(Ours) & \cellcolor{highlight}91.21 $\pm$ 6.89 & \cellcolor{highlight}86.99 & \cellcolor{highlight}86.48 & \cellcolor{highlight}77.07 \\
\bottomrule
\end{tabular}
\label{tab:Cholec80}
\vspace{-10pt}
\end{table*}

%% file: tables/GraSP.tex
\begin{table*}[t]
\aboverulesep=0ex
\belowrulesep=0ex
\setlength{\tabcolsep}{3pt} 
\renewcommand{\arraystretch}{1.15}
\centering
\small
\caption{Comparison of token merging methods on TAPIS (GRASP dataset). For the 16-layer backbone MViT, we apply temporal merging to Blocks 7-13. 
}
\begin{tabular}{c|c|c|c|c|cc|cc|c|c}
\toprule
\multirow{2}{*}{$R_T$} & \multirow{2}{*}{$R_S$} & \multirow{2}{*}{GFLOPs$\downarrow$} & \multirow{2}{*}{Memory(GB)$\downarrow$} & \multirow{2}{*}{Method} & \multicolumn{2}{c|}{Phases} & \multicolumn{2}{c|}{Steps} & Instruments & Actions \\ \cline{6-11} 
 &  &  &  &  & \multicolumn{1}{c|}{mAP$\uparrow$} & F1 score$\uparrow$ & \multicolumn{1}{c|}{mAP$\uparrow$} & F1 score$\uparrow$ & mAP@0.5IOU\_box$\uparrow$ & mAP$\uparrow$ \\ 
\midrule
- & - & 70.80 & 3.38 & Baseline & \multicolumn{1}{c|}{73.75} & 65.55 & \multicolumn{1}{c|}{51.38} & 47.01 & 86.59 & 39.50 \\ 
 1 & - & 47.98 & 2.69 & T-TESTA & \multicolumn{1}{c|}{68.73} & 59.17 & \multicolumn{1}{c|}{46.03} & 39.47 & 86.59 & 37.48 \\ 
 \rowcolor{highlight}
 1 & - & 47.98 & 2.69 & TIM-TM(Ours) & \multicolumn{1}{c|}{70.14} & 60.37 & \multicolumn{1}{c|}{46.39} & 40.24 & 86.56 & 38.13 \\ 
\bottomrule
\end{tabular}
\label{tab:GraSP}
\end{table*}

%% file: tables/foundationtable_mfig.tex
\begin{figure*}[htbp]
    \centering
    \begin{minipage}[t]{0.32\textwidth}
        \vspace{0pt}
        \aboverulesep=0ex
        \belowrulesep=0ex
        \setlength{\tabcolsep}{2pt}
        \renewcommand{\arraystretch}{1.0}
        \centering
        \small
        \captionof{table}{
             Comparison of merging methods on the EndoFM-LV classification task. Temporal (Blocks 1-6), Spatial (Blocks 7-12).
        }
        \begin{tabular}{cc|cc|c|c}
        \toprule
        $R_T$ & $R_S$ & GFLOPs$\downarrow$ & Mem.$\downarrow$ & Method & F1$\uparrow$ \\ \midrule
         - & - & 786.82 & 7.48 & Baseline & 97.5 \\
        \multirow{2}{*}{1} & \multirow{2}{*}{12} & 615.26 & 6.49 & ST-TESTA & 97.5  \\ 
         & & \cellcolor{highlight}615.21 & \cellcolor{highlight}6.78 & \cellcolor{highlight}STIM-TM & \cellcolor{highlight}97.5 \\ 
         \multirow{2}{*}{2} & \multirow{2}{*}{24} & 471.31 & 5.29 & ST-TESTA & 96.7  \\ 
         & & \cellcolor{highlight}471.28 & \cellcolor{highlight}5.74 & \cellcolor{highlight}STIM-TM & \cellcolor{highlight}97.5 \\ 
         \bottomrule
        \end{tabular}
        \label{tab:EndoFM-LV-classification}
    \end{minipage}
    \hfill
    \begin{minipage}[t]{0.28\textwidth}
        \vspace{0pt}
        \aboverulesep=0ex
        \belowrulesep=0ex
        \setlength{\tabcolsep}{2pt}
        \renewcommand{\arraystretch}{1.0}
        \centering
        \small
        \captionof{table}{
             Comparison of merging methods on the EndoFM-LV segmentation task. Temporal (Blocks 1-3), Spatial (Blocks 4-12). 
        }
        \begin{tabular}{cc|c|c|c}
        \toprule
        $R_T$ & $R_S$ & GFLOPs$\downarrow$ & Method & Dice$\uparrow$\\ \midrule
         - & - & 690.52 & Baseline & 83.6 \\
        \multirow{2}{*}{1} & \multirow{2}{*}{12} & 543.82 & ST-TESTA & 82.7\\ 
         & & \cellcolor{highlight}543.78 & \cellcolor{highlight}STIM-TM & \cellcolor{highlight}83.5 \\ 
         \multirow{2}{*}{1} & \multirow{2}{*}{24} & 467.69 & ST-TESTA & 82.5 \\ 
         & & \cellcolor{highlight}467.68 &\cellcolor{highlight}STIM-TM & \cellcolor{highlight}82.8 \\ 
         \bottomrule
        \end{tabular}
        \label{tab:EndoFM-LV-segmentation}
    \end{minipage}
    \hfill
    \begin{minipage}[t]{0.35\textwidth}
        \vspace{0pt}
        \centering
        \includegraphics[width=\textwidth]{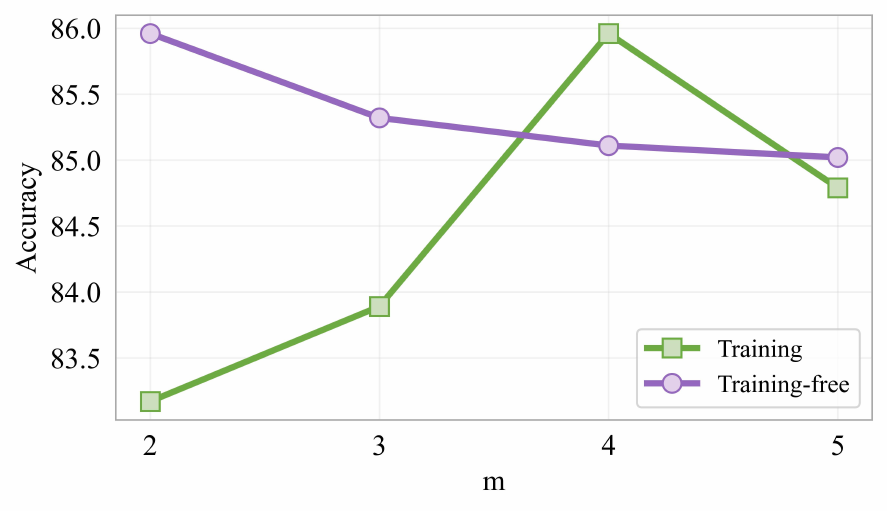}
        \vspace{-6mm}
        \caption{Performance comparison of different candidate size $m$ in training and training-free mode.}
        \label{fig:m_ablation}
    \end{minipage}
\end{figure*}

%% file: tables/IB.tex
\begin{table}[t]
\aboverulesep=0ex
\belowrulesep=0ex
\setlength{\tabcolsep}{1.2pt}
\renewcommand{\arraystretch}{1.0}
\centering
\caption{Comparison of IB scores for token reduction methods.}
\scriptsize
\resizebox{\linewidth}{!}{
\begin{tabular}{l|c|ccc}
    \toprule
    Method & Accuracy$\uparrow$ & $I(Z_{m}, X)\downarrow$ & $I(Z_{m}, Y)\uparrow$ & IB score$\downarrow$\\
    \midrule
    STTS~\cite{wang2022efficient} & 79.63 & \textbf{0.002057} & 0.209915  & -0.207858 \\
    STA~\cite{ding2023prune} & 80.30 & 0.002192 & 0.207727 & -0.205534 \\
    DynamicViT~\cite{rao2021dynamicvit} & 84.82 & 0.002814 & 0.270769 & -0.267955 \\
    ToMe~\cite{bolyatoken} & 80.99 & 0.003223 & 0.242318 & -0.239095 \\
    TempMe~\cite{shen2024tempme} & 84.09 & 0.002590 & 0.253951 & -0.251361 \\
    DyCoke~\cite{tao2024dycoke} & 81.64 & 0.002680 & 0.183614 & -0.180935 \\
    VisionZip~\cite{yang2025visionzip} & 84.52 & 0.003264 & 0.275990 & -0.272726 \\
    ST-TESTA~\cite{ren2023testa} & 85.23 & 0.003182 & 0.274063 & -0.270882 \\
    STIM-TM(Ours) & \textbf{86.04} & 0.002518 & \textbf{0.277328} & \textbf{-0.274810} \\
    \bottomrule
\end{tabular}
}
\label{tab:IB}
\end{table}

%% file: tables/sota_table_Autolaparo.tex
\begin{table}[t]
\aboverulesep=0ex
\belowrulesep=0ex
\setlength{\tabcolsep}{1.2pt}
\renewcommand{\arraystretch}{1.2}
\centering
\caption{Training with token merging methods. We plug ST-TESTA or STIM-TM into the Surgformer during both training and inference.}
\scriptsize
\resizebox{\linewidth}{!}{
\begin{tabular}{l|cc|cc|cccc}
    \toprule
    \multirow{2}{*}{Method} & \multirow{2}{*}{$R_T$} & \multirow{2}{*}{$R_S$} & \multirow{2}{*}{GFLOPs$\downarrow$} & \multirow{2}{*}{Mem.$\downarrow$} & Video-level& \multicolumn{3}{c}{Phase-level}\\
    \cmidrule(lr){6-6} \cmidrule(lr){7-9}
    & & & & & Acc.$\uparrow$ & Prec.$\uparrow$ & Rec.$\uparrow$ & Jacc.$\uparrow$ \\
    \midrule
    SV-RCNet~\cite{SV-RCNet} & - & - & - & - & 75.6 & 64.0 & 59.7 & 47.2 \\
    TMRNet~\cite{TMRNet} & - & - & - & - & 78.2 & 66.0 & 61.5 & 49.6 \\
    TeCNO~\cite{TeCNO} & - & - & - & - & 77.3 & 66.9 & 64.6 & 50.7 \\
    Trans-SVNet~\cite{Trans-SVNet} & - & - & - & - & 78.3 & 64.2 & 62.1 & 50.7 \\
    AVT~\cite{AVT} & - & - & - & - & 77.8 & 68.0 & 62.2 & 50.7 \\
    LoViT~\cite{LoViT} & - & - & - & - & 81.4$_{\scriptsize\pm 7.6}$ & 85.1 & 65.9 & 56.0 \\
    SKiT~\cite{SKiT} & - & - & - & - & 82.9$_{\scriptsize\pm 6.8}$ & 81.8 & 70.1 & 59.9 \\
    \midrule
    Surgformer~\cite{yang2024surgformer} & - & - & 918.78 & 21.90 & 85.9$_{\scriptsize\pm 5.8}$ & 83.4 & 74.3 & 65.1 \\
    \addlinespace[0.5pt]
    +T-TESTA & 1 & - & 562.04 & 15.65 & 83.4$_{\scriptsize\pm 7.7}$ & 77.6 & 72.2 & 62.5 \\
    +S-TESTA & - & 12 & 585.28 & 15.78 & 85.0$_{\scriptsize\pm 7.0}$ & 78.3 & 71.5 & 62.9 \\
    +ST-TESTA & 1 & 12 & 486.18 & 14.03 & 81.8$_{\scriptsize\pm 7.4}$ & 75.0 & 70.5 & 59.5 \\
    \addlinespace[0.5pt]
    \rowcolor{highlight}
    +TIM-TM(Ours) & 1 & - & 562.04 & 16.47 & 86.7$_{\scriptsize\pm 7.2}$ & 81.4 & 78.0 & 69.3 \\
    \rowcolor{highlight}
    +SIM-TM(Ours) & - & 12 & 599.22 & 14.88 & 86.0$_{\scriptsize\pm 4.5}$ & 81.8 & 74.7 & 65.8 \\
    \rowcolor{highlight}
    +STIM-TM(Ours) & 1 & 12 & 486.12 & 14.08 & 85.8$_{\scriptsize\pm 6.5}$ & 76.8 & 76.7 & 66.3 \\
    \bottomrule
\end{tabular}
}
\label{tab:r2}
\end{table}

%% file: tables/ablation_merging_rate.tex
\begin{table}[t!]
\aboverulesep=0ex
\belowrulesep=0ex
\setlength{\tabcolsep}{1.5pt}
\renewcommand{\arraystretch}{1.3}
\caption{Ablation study on merging rate. Results on Surgformer trained on Autolaparo dataset.}
\centering
\scriptsize
\resizebox{\linewidth}{!}{
\begin{tabular}{cc|cc|c|cccc}
    \toprule
    \multirow{2}{*}{$R_T$} & \multirow{2}{*}{$R_S$} & \multirow{2}{*}{GFLOPs$\downarrow$} & \multirow{2}{*}{Mem.$\downarrow$} & \multirow{2}{*}{Method} & Video-level& \multicolumn{3}{c}{Phase-level}  \\
    \cmidrule(lr){6-6} \cmidrule(lr){7-9}
    & & & & & Acc.$\uparrow$ & Prec.$\uparrow$ & Rec.$\uparrow$ & Jacc.$\uparrow$ \\
    \midrule
    - & - & 459.39 & 4.75 & Baseline & 85.9$_{\scriptsize\pm 5.8}$ & 83.4 & 74.3 & 65.1 \\  
    \addlinespace[0.5pt]
    \multirow{2}{*}{1} & \multirow{2}{*}{12} & 300.40 & 3.74 & ST-TESTA & 85.2$_{\scriptsize\pm 6.1}$ & 78.4 & 74.6 & 64.7 \\
    & & \cellcolor{highlight}300.37 & \cellcolor{highlight}3.77 & \cellcolor{highlight}STIM-TM & \cellcolor{highlight}86.0$_{\scriptsize\pm 5.6}$ & \cellcolor{highlight}80.5 & \cellcolor{highlight}74.3 & \cellcolor{highlight}65.2 \\
    \addlinespace[0.5pt]
    \multirow{2}{*}{2} & \multirow{2}{*}{12} & 187.79 & 2.84 & ST-TESTA & 76.0$_{\scriptsize\pm 7.1}$ & 66.8 & 69.0 & 52.6 \\
    & & \cellcolor{highlight}187.77 & \cellcolor{highlight}2.94 & \cellcolor{highlight}STIM-TM & \cellcolor{highlight}82.6$_{\scriptsize\pm 5.6}$ & \cellcolor{highlight}72.9 & \cellcolor{highlight}71.6 & \cellcolor{highlight}60.0 \\
    \addlinespace[0.5pt]
    \multirow{2}{*}{2} & \multirow{2}{*}{24} & 177.62 & 2.78 & ST-TESTA & 75.7$_{\scriptsize\pm 7.0}$ & 67.3 & 69.2 & 52.7 \\
    & & \cellcolor{highlight}177.61 & \cellcolor{highlight}2.86 & \cellcolor{highlight}STIM-TM & \cellcolor{highlight}82.6$_{\scriptsize\pm 5.4}$ & \cellcolor{highlight}73.9 & \cellcolor{highlight}71.9 & \cellcolor{highlight}60.3 \\
    \bottomrule
\end{tabular}
}
\label{tab:ablation_mergingrate}
\end{table}

%% file: sections/5_discussion.tex
\section{Conclusion}
\label{sec:discussion}
This work presents STIM-TM, the first dedicated token merging framework for surgical video understanding. Leveraging information-theoretic insights into video redundancy patterns, our method introduces a decoupled approach that independently addresses temporal and spatial redundancy while preserving task-relevant information. Extensive validation demonstrates STIM-TM's effectiveness, achieving substantial computational efficiency gains with minimal performance degradation during training and inference. As a training-free, plug-and-play solution, STIM-TM enables practical deployment of vision transformers in resource-constrained surgical environments, advancing the development of intelligent surgical systems for real-world clinical applications.

%% file: tmi.bbl
\begin{thebibliography}{10}
\providecommand{\url}[1]{#1}
\csname url@samestyle\endcsname
\providecommand{\newblock}{\relax}
\providecommand{\bibinfo}[2]{#2}
\providecommand{\BIBentrySTDinterwordspacing}{\spaceskip=0pt\relax}
\providecommand{\BIBentryALTinterwordstretchfactor}{4}
\providecommand{\BIBentryALTinterwordspacing}{\spaceskip=\fontdimen2\font plus
\BIBentryALTinterwordstretchfactor\fontdimen3\font minus \fontdimen4\font\relax}
\providecommand{\BIBforeignlanguage}[2]{{%
\expandafter\ifx\csname l@#1\endcsname\relax
\typeout{** WARNING: IEEEtran.bst: No hyphenation pattern has been}%
\typeout{** loaded for the language `#1'. Using the pattern for}%
\typeout{** the default language instead.}%
\else
\language=\csname l@#1\endcsname
\fi
#2}}
\providecommand{\BIBdecl}{\relax}
\BIBdecl

\bibitem{khan2025surgical}
U.~Khan, U.~Nawaz, A.~Qayyum, S.~Ashraf, M.~Bilal, and J.~Qadir, ``Surgical scene understanding in the era of foundation ai models: A comprehensive review,'' \emph{arXiv preprint arXiv:2502.14886}, 2025.

\bibitem{LoViT}
Y.~Liu, M.~Boels, L.~C. Garcia-Peraza-Herrera, T.~Vercauteren, P.~Dasgupta, A.~Granados, and S.~Ourselin, ``Lovit: Long video transformer for surgical phase recognition,'' \emph{Medical Image Analysis}, vol.~99, p. 103366, 2025.

\bibitem{yang2024surgformer}
S.~Yang, L.~Luo, Q.~Wang, and H.~Chen, ``Surgformer: Surgical transformer with hierarchical temporal attention for surgical phase recognition,'' in \emph{International Conference on Medical Image Computing and Computer-Assisted Intervention}.\hskip 1em plus 0.5em minus 0.4em\relax Springer, 2024, pp. 606--616.

\bibitem{ayobi2024pixel}
N.~Ayobi, S.~Rodr{\'\i}guez, A.~P{\'e}rez, I.~Hern{\'a}ndez, N.~Aparicio, E.~Dessevres, S.~Pe{\~n}a, J.~Santander, J.~I. Caicedo, N.~Fern{\'a}ndez \emph{et~al.}, ``Pixel-wise recognition for holistic surgical scene understanding,'' \emph{arXiv preprint arXiv:2401.11174}, 2024.

\bibitem{liu2025resurgsam2}
H.~Liu, M.~Gao, X.~Luo, Z.~Wang, G.~Qin, J.~Wu, and Y.~Jin, ``Resurgsam2: Referring segment anything in surgical video via credible long-term tracking,'' \emph{International Conference on Medical Image Computing and Computer Assisted Intervention}, 2025.

\bibitem{pei2025instrument}
J.~Pei, J.~Zhang, G.~Qin, K.~Wang, Y.~Jin, and P.-A. Heng, ``Instrument-tissue-guided surgical action triplet detection via textual-temporal trail exploration,'' \emph{IEEE Transactions on Medical Imaging}, 2025.

\bibitem{dosovitskiy2020image}
A.~Dosovitskiy, L.~Beyer, A.~Kolesnikov, D.~Weissenborn, X.~Zhai, T.~Unterthiner, M.~Dehghani, M.~Minderer, G.~Heigold, S.~Gelly \emph{et~al.}, ``An image is worth 16x16 words: Transformers for image recognition at scale,'' \emph{International Conference on Learning Representations}, 2021.

\bibitem{bolyatoken}
D.~Bolya, C.-Y. Fu, X.~Dai, P.~Zhang, C.~Feichtenhofer, and J.~Hoffman, ``Token merging: Your vit but faster,'' in \emph{The Eleventh International Conference on Learning Representations}, 2023.

\bibitem{wang2025improving}
Z.~Wang, C.~Liu, L.~Zhu, T.~Wang, S.~Zhang, and Q.~Dou, ``Improving foundation model for endoscopy video analysis via representation learning on long sequences,'' \emph{IEEE Journal of Biomedical and Health Informatics}, 2025.

\bibitem{liu2022videoswin}
Z.~Liu, J.~Ning, Y.~Cao, Y.~Wei, Z.~Zhang, S.~Lin, and H.~Hu, ``Video swin transformer,'' in \emph{Proceedings of the IEEE/CVF conference on computer vision and pattern recognition}, 2022, pp. 3202--3211.

\bibitem{timesformer21}
G.~Bertasius, H.~Wang, and L.~Torresani, ``Is space-time attention all you need for video understanding?'' in \emph{International Conference on Machine Learning}, vol.~2, no.~3, 2021, p.~4.

\bibitem{SKiT}
Y.~Liu, J.~Huo, J.~Peng, R.~Sparks, P.~Dasgupta, A.~Granados, and S.~Ourselin, ``Skit: a fast key information video transformer for online surgical phase recognition,'' in \emph{Proceedings of the IEEE/CVF International Conference on Computer Vision}, 2023, pp. 21\,074--21\,084.

\bibitem{norouzi2024algm}
N.~Norouzi, S.~Orlova, D.~De~Geus, and G.~Dubbelman, ``Algm: Adaptive local-then-global token merging for efficient semantic segmentation with plain vision transformers,'' in \emph{Proceedings of the IEEE/CVF Conference on Computer Vision and Pattern Recognition}, 2024, pp. 15\,773--15\,782.

\bibitem{lee2024multi}
S.~Lee, J.~Choi, and H.~J. Kim, ``Multi-criteria token fusion with one-step-ahead attention for efficient vision transformers,'' in \emph{Proceedings of the IEEE/CVF Conference on Computer Vision and Pattern Recognition}, 2024, pp. 15\,741--15\,750.

\bibitem{yang2025visionzip}
S.~Yang, Y.~Chen, Z.~Tian, C.~Wang, J.~Li, B.~Yu, and J.~Jia, ``Visionzip: Longer is better but not necessary in vision language models,'' in \emph{Proceedings of the Computer Vision and Pattern Recognition Conference}, 2025, pp. 19\,792--19\,802.

\bibitem{shang2025prumerge}
Y.~Shang, M.~Cai, B.~Xu, Y.~J. Lee, and Y.~Yan, ``Llava-prumerge: Adaptive token reduction for efficient large multimodal models,'' in \emph{ICCV}, 2025.

\bibitem{tao2024dycoke}
K.~Tao, C.~Qin, H.~You, Y.~Sui, and H.~Wang, ``Dycoke: Dynamic compression of tokens for fast video large language models,'' \emph{The IEEE/CVF Conference on Computer Vision and Pattern Recognition}, 2025.

\bibitem{choi2024vid}
J.~Choi, S.~Lee, J.~Chu, M.~Choi, and H.~J. Kim, ``vid-tldr: Training free token merging for light-weight video transformer,'' in \emph{Proceedings of the IEEE/CVF Conference on Computer Vision and Pattern Recognition}, 2024, pp. 18\,771--18\,781.

\bibitem{lee2024video}
S.-H. Lee, J.~Wang, Z.~Zhang, D.~Fan, and X.~Li, ``Video token merging for long video understanding,'' \emph{Advances in Neural Information Processing Systems}, vol.~37, pp. 13\,851--13\,871, 2024.

\bibitem{shen2024tempme}
L.~Shen, T.~Hao, T.~He, S.~Zhao, Y.~Zhang, P.~Liu, Y.~Bao, and G.~Ding, ``Tempme: Video temporal token merging for efficient text-video retrieval,'' \emph{The Thirteenth International Conference on Learning Representations}, 2025.

\bibitem{wang2023foundation}
Z.~Wang, C.~Liu, S.~Zhang, and Q.~Dou, ``Foundation model for endoscopy video analysis via large-scale self-supervised pre-train,'' in \emph{International Conference on Medical Image Computing and Computer-Assisted Intervention}.\hskip 1em plus 0.5em minus 0.4em\relax Springer, 2023, pp. 101--111.

\bibitem{jaspers2025scaling}
T.~J. Jaspers, R.~L. de~Jong, Y.~Li, C.~H. Kusters, F.~H. Bakker, R.~C. van Jaarsveld, G.~M. Kuiper, R.~van Hillegersberg, J.~P. Ruurda, W.~M. Brinkman \emph{et~al.}, ``Scaling up self-supervised learning for improved surgical foundation models,'' \emph{arXiv preprint arXiv:2501.09436}, 2025.

\bibitem{rao2021dynamicvit}
Y.~Rao, W.~Zhao, B.~Liu, J.~Lu, J.~Zhou, and C.-J. Hsieh, ``Dynamicvit: Efficient vision transformers with dynamic token sparsification,'' \emph{Advances in neural information processing systems}, vol.~34, pp. 13\,937--13\,949, 2021.

\bibitem{wang2022efficient}
J.~Wang, X.~Yang, H.~Li, L.~Liu, Z.~Wu, and Y.-G. Jiang, ``Efficient video transformers with spatial-temporal token selection,'' in \emph{European Conference on Computer Vision}.\hskip 1em plus 0.5em minus 0.4em\relax Springer, 2022, pp. 69--86.

\bibitem{ding2023prune}
S.~Ding, P.~Zhao, X.~Zhang, R.~Qian, H.~Xiong, and Q.~Tian, ``Prune spatio-temporal tokens by semantic-aware temporal accumulation,'' in \emph{Proceedings of the IEEE/CVF International Conference on Computer Vision}, 2023, pp. 16\,945--16\,956.

\bibitem{ren2023testa}
S.~Ren, S.~Chen, S.~Li, X.~Sun, and L.~Hou, ``Testa: Temporal-spatial token aggregation for long-form video-language understanding,'' in \emph{The 2023 Conference on Empirical Methods in Natural Language Processing}, 2023.

\bibitem{li2024vidtome}
X.~Li, C.~Ma, X.~Yang, and M.-H. Yang, ``Vidtome: Video token merging for zero-shot video editing,'' in \emph{Proceedings of the IEEE/CVF Conference on Computer Vision and Pattern Recognition}, 2024, pp. 7486--7495.

\bibitem{su2024stpm}
Y.~Su, J.~Zhang, R.~Yan, P.~Li, G.~Xie, and X.~Shu, ``Stpm: Spatial-temporal token pruning and merging for complex activity recognition,'' \emph{IEEE Transactions on Circuits and Systems for Video Technology}, 2024.

\bibitem{wang2022autolaparo}
Z.~Wang, B.~Lu, Y.~Long, F.~Zhong, T.-H. Cheung, Q.~Dou, and Y.~Liu, ``Autolaparo: A new dataset of integrated multi-tasks for image-guided surgical automation in laparoscopic hysterectomy,'' in \emph{International Conference on Medical Image Computing and Computer-Assisted Intervention}.\hskip 1em plus 0.5em minus 0.4em\relax Springer, 2022, pp. 486--496.

\bibitem{hong2020cholecseg8k}
W.-Y. Hong, C.-L. Kao, Y.-H. Kuo, J.-R. Wang, W.-L. Chang, and C.-S. Shih, ``Cholecseg8k: a semantic segmentation dataset for laparoscopic cholecystectomy based on cholec80,'' \emph{arXiv preprint arXiv:2012.12453}, 2020.

\bibitem{oord2018representation}
A.~v.~d. Oord, Y.~Li, and O.~Vinyals, ``Representation learning with contrastive predictive coding,'' \emph{arXiv preprint arXiv:1807.03748}, 2018.

\bibitem{tishby2015deep}
N.~Tishby and N.~Zaslavsky, ``Deep learning and the information bottleneck principle,'' in \emph{2015 ieee information theory workshop (itw)}.\hskip 1em plus 0.5em minus 0.4em\relax Ieee, 2015, pp. 1--5.

\bibitem{wang2025efficient}
Y.~Wang and Y.~Yang, ``Efficient visual transformer by learnable token merging,'' \emph{IEEE Transactions on Pattern Analysis and Machine Intelligence}, 2025.

\bibitem{wang2024videollamb}
Y.~Wang, C.~Xie, Y.~Liu, and Z.~Zheng, ``Videollamb: Long-context video understanding with recurrent memory bridges,'' \emph{arXiv preprint arXiv:2409.01071}, 2024.

\bibitem{twinanda2016endonet}
A.~P. Twinanda, S.~Shehata, D.~Mutter, J.~Marescaux, M.~De~Mathelin, and N.~Padoy, ``Endonet: a deep architecture for recognition tasks on laparoscopic videos,'' \emph{IEEE transactions on medical imaging}, vol.~36, no.~1, pp. 86--97, 2016.

\bibitem{fan2021multiscale}
H.~Fan, B.~Xiong, K.~Mangalam, Y.~Li, Z.~Yan, J.~Malik, and C.~Feichtenhofer, ``Multiscale vision transformers,'' in \emph{Proceedings of the IEEE/CVF International Conference on Computer Vision}, 2021, pp. 6824--6835.

\bibitem{cheng2022masked}
B.~Cheng, I.~Misra, A.~G. Schwing, A.~Kirillov, and R.~Girdhar, ``Masked-attention mask transformer for universal image segmentation,'' in \emph{Proceedings of the IEEE/CVF conference on computer vision and pattern recognition}, 2022, pp. 1290--1299.

\bibitem{tian2022contrastive}
Y.~Tian, G.~Pang, F.~Liu, Y.~Liu, C.~Wang, Y.~Chen, J.~Verjans, and G.~Carneiro, ``Contrastive transformer-based multiple instance learning for weakly supervised polyp frame detection,'' in \emph{International Conference on Medical Image Computing and Computer-Assisted Intervention}.\hskip 1em plus 0.5em minus 0.4em\relax Springer, 2022, pp. 88--98.

\bibitem{bernal2015wm}
J.~Bernal, F.~J. S{\'a}nchez, G.~Fern{\'a}ndez-Esparrach, D.~Gil, C.~Rodr{\'\i}guez, and F.~Vilari{\~n}o, ``Wm-dova maps for accurate polyp highlighting in colonoscopy: Validation vs. saliency maps from physicians,'' \emph{Computerized medical imaging and graphics}, vol.~43, pp. 99--111, 2015.

\bibitem{SV-RCNet}
Y.~Jin, Q.~Dou, H.~Chen, L.~Yu, J.~Qin, C.-W. Fu, and P.-A. Heng, ``Sv-rcnet: workflow recognition from surgical videos using recurrent convolutional network,'' \emph{IEEE transactions on medical imaging}, vol.~37, no.~5, pp. 1114--1126, 2017.

\bibitem{TMRNet}
Y.~Jin, Y.~Long, C.~Chen, Z.~Zhao, Q.~Dou, and P.-A. Heng, ``Temporal memory relation network for workflow recognition from surgical video,'' \emph{IEEE Transactions on Medical Imaging}, vol.~40, no.~7, pp. 1911--1923, 2021.

\bibitem{TeCNO}
T.~Czempiel, M.~Paschali, M.~Keicher, W.~Simson, H.~Feussner, S.~T. Kim, and N.~Navab, ``Tecno: Surgical phase recognition with multi-stage temporal convolutional networks,'' in \emph{Medical Image Computing and Computer Assisted Intervention--MICCAI 2020: 23rd International Conference, Lima, Peru, October 4--8, 2020, Proceedings, Part III 23}.\hskip 1em plus 0.5em minus 0.4em\relax Springer, 2020, pp. 343--352.

\bibitem{Trans-SVNet}
X.~Gao, Y.~Jin, Y.~Long, Q.~Dou, and P.-A. Heng, ``Trans-svnet: Accurate phase recognition from surgical videos via hybrid embedding aggregation transformer,'' in \emph{Medical Image Computing and Computer Assisted Intervention--MICCAI 2021: 24th International Conference, Strasbourg, France, September 27--October 1, 2021, Proceedings, Part IV 24}.\hskip 1em plus 0.5em minus 0.4em\relax Springer, 2021, pp. 593--603.

\bibitem{AVT}
R.~Girdhar and K.~Grauman, ``Anticipative video transformer,'' in \emph{Proceedings of the IEEE/CVF international conference on computer vision}, 2021, pp. 13\,505--13\,515.

\end{thebibliography}
